\documentclass[letterpaper, 10pt, conference]{IEEEtran}
\IEEEoverridecommandlockouts
\def\IEEEtitletopspace{18pt}
\usepackage{cite}
\usepackage{amsmath,amssymb,amsfonts}
\usepackage{algorithmic}
\usepackage{graphicx}
\usepackage{textcomp}
\usepackage{xcolor}
\usepackage{hyperref}
\usepackage{float}

\usepackage{enumerate}
\usepackage{romannum}
\usepackage{multirow}
\usepackage{wrapfig}

\usepackage{hyperref}
\hypersetup{hidelinks}
\usepackage[caption=false,font=footnotesize]{subfig}
\usepackage{makecell}
\usepackage{ragged2e}
\usepackage[ruled,vlined]{algorithm2e}
\usepackage{algorithmic}

\def\BibTeX{{\rm B\kern-.05em{\sc i\kern-.025em b}\kern-.08em
    T\kern-.1667em\lower.7ex\hbox{E}\kern-.125emX}}
\begin{document}

\title{Suturing Tasks Automation Based on Skills Learned From Demonstrations: A Simulation Study
}

\author{Haoying Zhou$^{1,3}$, Yiwei Jiang$^{1}$, Shang Gao$^{1}$, Shiyue Wang$^{3}$, Peter Kazanzides$^{2,3}$ and Gregory S. Fischer$^{1}$

\thanks{This work was supported in part by NSF AccelNet award OISE-1927275 and OISE-192735.}
\thanks{$^{1}$Department of Robotics Engineering, Worcester Polytechnic Institute, Worcester, MA, USA. (Emails: {hzhou6, yjiang5, sgao, gfischer}@wpi.edu)}
\thanks{$^{2}$Department of Computer Science, Johns Hopkins University, Baltimore, MD, USA. (Email: {pkaz}@jhu.edu)}
\thanks{$^{3}$Laboratory for Computational Sensing and Robotics, Johns Hopkins University, Baltimore, MD, USA.}
}



\maketitle


\begin{abstract}
In this work, we develop an open-source surgical simulation environment that includes a realistic model obtained by MRI-scanning a physical phantom, for the purpose of training and evaluating a Learning from Demonstration (LfD) algorithm for autonomous suturing. The LfD algorithm utilizes Dynamic Movement Primitives (DMP) and Locally Weighted Regression (LWR), but focuses on the needle trajectory, rather than the instruments, to obtain better generality with respect to needle grasps. We conduct a user study to collect multiple suturing demonstrations and perform a comprehensive analysis of the ability of the LfD algorithm to generalize from a demonstration at one location in one phantom to different locations in the same phantom and to a different phantom. Our results indicate good generalization, on the order of 91.5\%, when learning from more experienced subjects, indicating the need to integrate skill assessment in the future.
\end{abstract}



\section{Introduction}

Robotic surgery has revolutionized the field of medical science by providing surgeons with enhanced dexterity, visualization, and precision during minimally invasive procedures. In the recent two decades, the usage of surgical robots in hospitals has significantly increased \cite{sheetz2020trends}. As the technology becomes more widespread, there is an increasing need to improve task autonomy \cite{attanasio2021autonomy} for surgical robots to facilitate their use by clinicians.

For instance, the da Vinci{\textregistered} Surgical System (dVSS), one of the most widely adopted robotic platforms \cite{d2004robotic}, has transformed surgical interventions, enabling complex procedures with improved patient outcomes. However, despite the remarkable capabilities of robotic systems, certain repetitive tasks, such as suturing, continue to demand significant manual intervention, which can impose cognitive load on surgeons and impact procedural efficiency.

Automating repetitive surgical tasks has emerged as a promising approach to alleviate the cognitive burden on surgeons, allowing them to focus on critical decision-making and enhancing patient care. In this regard, automating suturing has gained significant attention, aiming to reduce surgical time, enhance precision, and ensure consistent suture placement, thereby optimizing surgical outcomes.

In this paper, we present a method for automatic suturing path planning using the da Vinci Research Kit (dVRK) \cite{kazanzides2014open} in simulation with comprehensive analysis. Taking advantage of the Asynchronous Multi-Body Framework (AMBF) simulator \cite{munawar2019real, munawar2022open} that integrates seamlessly with dVRK and adding a phantom volume scanned by MRI, we construct a realistic simulation surgical scene to perform the suturing procedures for training the automation algorithms. Our approach leverages Learning from Demonstration (LfD) with Dynamic Movement Primitives (DMPs) \cite{Ijspeert2002, Schaal2006, ijspeert2013dynamical} and Locally Weighted Regression (LWR) \cite{Schaal1998}. Also, we construct a user study to collect data from human subjects to train the robot on performing suturing tasks. Notably, this simulation scene allows us to have much higher flexibility for phantom and environment selections and overcome the restrictions on demonstration collection. Moreover, the simulation enables us to obtain the ground truth of objects for motion analysis and generality tests. Furthermore, the trained model is anticipated to be directly deployed to the physical dVRK, ensuring its practical applicability.

In summary, the contributions of this work are:

\begin{itemize}
    \item A novel and realistic simulation environment using an MRI-scanned phantom
    \item A recording pipeline for suturing automation data collection in the simulation
    \item A method to improve the generality of the LfD algorithm by selecting the suture needle as the learning object 
    \item A comprehensive assessment of generality by testing at different positions and with different phantoms
\end{itemize}


\section{Related Works}

A series of researches have been done to push forward the effort on real-world bi-manual suturing tasks automation. Sen et al.\cite{sen2016automating} proposed the Suture Needle Angular Positioner (SNAP) to ensure a constant transformation between the robot manipulator and the suture needle so that it can facilitate the suturing automation procedure with sequential convex programming. The known transformation not only enables the researchers to regard the needle as an extended end-effector for calculating the kinematics, but also excludes the variance due to different needle grasps. On the other hand, the constant transformation restricts the generality of the algorithm. Varier et al. \cite{varier2020collaborative} first introduced Reinforcement Learning (RL) to accomplish the suturing automation. Nevertheless, Varier's method requires to discretize the workspace of the robot which is a challenging task when implementing in real surgeries. Also, depending on the grid size, it might need tremendous computational power to enable the algorithm. Schwaner et al. \cite{schwaner2021autonomous_needle, schwaner2021autonomous} presented a LfD algorithm prototype of suturing automation with DMPs using UR5 robot arms holding a da Vinci Large Needle Driver. Schwaner's method integrates with computer-vision-based tracking and implements the algorithm on a real-world robot. Nevertheless, it collects demonstrations from the real world, which can be a challenging and time-consuming process. In addition, the complexity of collecting real-life demonstrations leads to greater difficulty when performing generality tests.


Due to the restrictions in the real world and ethical concerns, simulation environments have been developed to facilitate the process of obtaining demonstrations for LfD and medical training\cite{munawar2019real, munawar2022open, allard2007sofa, lungu2021review,  haiderbhai2022robust}. Collecting demonstrations from simulation can overcome some difficulties of obtaining ground truth when performing suturing tasks, such as suture needle tracking\cite{jiang2023markerless}, and allows experimentation without risk. An ideal simulation can also optimize the experiment design by eliminating the random error due to the noise from the robot or the external environment and help to construct a better understanding of human motion patterns.


\section{Methodology}

The following two sections present standard formulations of Dynamic Movement Primitives (DMP) and Locally Weighted Regression (LWR) as background information. These are followed by Section \ref{sec:pipeline}, which describes the LfD pipeline that is based on the needle trajectory.

\subsection{Dynamic Movement Primitives}
\label{sec:DMP}

DMP\cite{Ijspeert2002, Schaal2006, ijspeert2013dynamical, ijspeert2003learning, schaal2003control, Pervez2018, ginesi2020autonomous} is a method for trajectory control and planning, which can represent complex motor actions without manual parameter tuning. In this work, we use discrete DMP to learn a point-to-point trajectory from a demonstration. Corresponding methodologies for position and orientation regeneration are shown in the following sections.

\subsubsection{Position}

We utilize DMPs to encode Cartesian space robot position trajectories and generate the learning weights. DMPs can represent a movement trajectory with a group of second-order ordinary differential equations, as shown in \autoref{eq: dmp_pos} \cite{ijspeert2013dynamical}:
\begin{equation}
    \begin{split}
        \tau^2 \Ddot{y} &= \alpha_{y}(\beta_{y}(g-y) + \tau\Dot{y}) + f(x)\\
        \tau \Dot{x} &= - \alpha_{x} x
    \end{split}
\label{eq: dmp_pos}
\end{equation}
where $y$, which can also be represented as $y(t) \in \mathbb{R}^3$, is the Cartesian space position of the robot system's end effector at time $t$; $x$ is a system variable which initiates at 1 and diminishes to 0 over time; $\alpha_x$, $\alpha_y$ and $\beta_y$ $\in \mathbb{R}^+$ are the gain coefficients; $g \in \mathbb{R}^3$ is the goal position of the end effector; $\tau \in \mathbb{R}^+$ is the time constant; and $f(x)$ is the nonlinear forcing term represented by \autoref{eq: dmp_pos_force}.  Regardless of how the gain coefficients are chosen, the robot system will always converge to $g$ because the influence of $f(x)$ will be ignored when $x \rightarrow 0$. This feature can ensure the stability of the system.

The nonlinear forcing term $f(x)$ can be represented as\cite{ijspeert2013dynamical}:
\begin{equation}
    \begin{split}
        f(x) &= \frac{\sum_{i=1}^{N_{bfs}}\psi_{i}(x)w_{i}}{\sum_{i=1}^{N_{bfs}}\psi_{i}(x)}x(g-y_0) \\
        \psi_i(x) &= e^{-h_i (x - c_i)^2} \\
        h_i &= \frac{{N_{bfs}}^{1.5}}{\alpha_x c_i}
    \end{split}
\label{eq: dmp_pos_force}
\end{equation}
where $\psi_i(x)$ is a basis function, which is a Gaussian function; $N_{bfs}$ is the number of basis functions; $y_0 \in \mathbb{R}^3$ is the initial position of the end effector; $w_i \in \mathbb{R}^3$ represents the learning weights; $c_i$ is the center of the Gaussian function and $h_i$ is the variance of the Gaussian function.

For learning from demonstration, we can calculate the desired nonlinear term $f_{des}(x)$ from a given demonstration trajectory $[y_{demo}, \Dot{y}_{demo}, \Ddot{y}_{demo}]$, and then we can obtain the learning weights $w_i$ to regenerate the trajectories:
\begin{equation}
    \begin{split}
        f_{des}(x) &= \tau^2 \Ddot{y}_{demo} - \alpha_y ( \beta_y (g_{demo} - y_{demo}) + \tau \Dot{y}_{demo})  \\
        &= \frac{\sum_{i=1}^{N_{bfs}}\psi_{i}(x)w_{i}}{\sum_{i=1}^{N_{bfs}}\psi_{i}(x)}x(g_{demo}-{y_{demo}}(0)) 
    \end{split}
\label{eq: dmp_forcing_pos}
\end{equation}
where $y_{demo}(0)$ is the initial state of the demonstration trajectory and $g_{demo}$ is the goal state of the demonstration trajectory.

For the time constant $\tau$, we can move it into the gain coefficients $\alpha_x$, $\beta_y$ and $\alpha_y$. Therefore, without losing any generality, we can set the time constant $\tau = 1$ for simplification. After calculating the learning weights via LWR and substituting the learning weights into \autoref{eq: dmp_pos}, we can obtain the $\Ddot{y}$ in \autoref{eq: dmp_pos} and integrate $\Ddot{y}$ to regenerate the trajectory $y$.

\subsubsection{Orientation}

For orientation, we utilize quaternions \cite{chiaverini1999unit, ude1999filtering, sabatini2006quaternion, faraway2009modelling}, $\mathbf{q} = v + \mathbf{u} \in S^3 $, to represent the angular movements, where $v \in \mathbb{R} $, $\mathbf{u} \in \mathbb{R}^3$ and $S^3$ is a unit sphere in $\mathbb{R}^4$. For this kind of representation, $q$ and $-q$ are the same orientation. Similar to \autoref{eq: dmp_pos},  DMPs can also represent an angular movement trajectory with a group of second-order ordinary differential equations as follows\cite{Gams}:
\begin{equation}
    \begin{split}
        \tau^2 \Dot{\mathbf{\omega}} &= \alpha_{z}(\beta_{z}2log(g_o*\overline{q}) - \tau\mathbf{\omega}) + f_o(x)\\
        \Dot{q} &= \frac{1}{2} \mathbf{\omega} * q \\
        \mathbf{\omega} &= 2log(g_o*\overline{q})
    \end{split}
\label{eq: dmp_ori}
\end{equation}
where $log(\cdot)$ refers to the natural logarithm\cite{ude1999filtering}; $\overline{q}$ is the quaternion conjugation; $g_o \in S^3$ is the goal quaternion orientation; $\mathbf{\omega} \in \mathbb{R}^3$, which can also be denoted as $\mathbf{\omega}(t)$, is the angular velocity of the robot system at time $t$; $x$ is a system variable identical to \autoref{eq: dmp_pos}; $\alpha_z$ and $\beta_z$ are the gain coefficients; $\tau$ is the time constant; $f_o(x)$ is the nonlinear forcing term. As in the previous section, we can set the time constant $\tau = 1$ for simplification. The rules of operations for quaternions are defined in the papers\cite{sabatini2006quaternion, faraway2009modelling}.


The nonlinear forcing term $f_o(x)$ can be represented as:
\begin{equation}
    \begin{split}
        f_o(x) &= D_o \frac{\sum_{i=1}^{N_{bfs}^o}\psi_{i}(x)w_{i}^o}{\sum_{i=1}^{N_{bfs}^o}\psi_{i}(x)}x 
    \end{split}
\label{eq: dmp_ori_force}
\end{equation}
where $D_o = diag(2log(g_o*q_0)) \in \mathbb{R}^{3 \times 3}$ is the scaling term; $q_0$ is the initial state of the orientation trajectory; $\psi_i$ is the basis function shown in \autoref{eq: dmp_pos_force}; $w_i^o \in \mathbb{R}^3$ is the learning weight for orientation trajectories.


Similar to \autoref{eq: dmp_forcing_pos} , given the demonstration orientation trajectory $[q_{des}, \mathbf{\omega}_{des}, \Dot{\mathbf{\omega}}_{des}]$ and \autoref{eq: dmp_ori}, we can calculate the learning weights $\mathbf{w}_i^o$ via solving the following equation:
\begin{equation}
\footnotesize
    \begin{split}
        \frac{\sum_{i=1}^{N_{bfs}^o}\psi_{i}(x)w_{i}^o}{\sum_{i=1}^{N_{bfs}^o}\psi_{i}(x)}x 
        =  D_o^{-1} (\Dot{\mathbf{\omega}}_{des} - \alpha_z (\beta_z 2log(g_{des}^o*\overline{q}_{des}) - \mathbf{\omega}_{des}))
    \end{split}
    \label{eq:dmp_forcing_ori}
\end{equation}
where $g_{des}^o$ is the goal state of the demonstration orientation trajectory. 

After obtaining the learning weights, we can integrate the quaternions in \autoref{eq: dmp_ori} using the following formula:
\begin{equation}
    \begin{split}
        \mathbf{q}(t +\Delta t) &= e^{\frac{\mathbf{\omega}(t) \Delta t }{2}}*\mathbf{q}(t) \\
        e^{\mathbf{r}} &= \left\{ \begin{array}{lr} cos(\lVert \mathbf{r} \rVert) + sin(\lVert \mathbf{r} \rVert) \frac{\mathbf{r}}{\lVert \mathbf{r} \rVert} & \mathbf{r} \neq \mathbf{0} \\ \mathbf{0} & \textrm{otherwise} \end{array}\right.
    \end{split}
    \label{eq:integrate_q}
\end{equation}

If we limit the domain of the exponential map $e^{\mathbf{r}}$ : $\mathbb{R}^3 \rightarrow S^3$ to $\lVert \mathbf{r} \rVert < \pi$ and the domain of the logarithmic map to $S^3 / (-1+[0,0,0]^T)$, then both mappings become one-to-one, continuously differentiable and inverse to each other. In addition, we also utilize phase stopping and goal switching\cite{Gams} techniques for better performance.  


\subsection{Locally Weighted Regression}
\label{sec:LWR}

Locally weighted regression\cite{Schaal1998} belongs to a class of nonparametric statistical techniques called locally weighted learning (LWL). LWR is a memory-based learning algorithm that can efficiently represent and train complex motor movements in autonomous adaptive control of robotic systems. The key advantage of LWR is its fast training speed, which only requires adding new training data to the memory.

We can use LWR to calculate the optimal learning weights $w_i$ for positions in \autoref{eq: dmp_forcing_pos}. Then, the cost function to be minimized is defined as:
\begin{equation}
    \begin{split}
        J(w_i) = \sum\psi_i(x)(f_{des}(x)-w_i(x(g_{demo}-y_{demo}(0))))^2
    \end{split}
    \label{eq:LWR_cost}
\end{equation}

Since $x$, which can be also denoted as $x(t)$, is a function of time $t$, therefore, denoting $y_{offset} = g_{demo}-y_{demo}(0)$,  we can rewrite \autoref{eq:LWR_cost} as follows:
\begin{equation}
    \begin{split}
        J(w_i) = \sum\psi_i(t)(f_{des}(t)-w_i(x(t)y_{offset})^2
    \end{split}
    \label{eq:LWR_cost_t}
\end{equation}

For solving \autoref{eq:LWR_cost_t}, construct a diagonal matrix $\Psi_i$ using $\psi_i$ along with time:
\begin{equation}
\small
    \begin{split}
        \Psi_{i}=\begin{bmatrix}
        \psi_{i}(t_{0}) & \cdots & 0 & 0\\
        \vdots & \psi_{i}(t_{1}) &\ddots &0\\
        0 &\ddots&\ddots&\vdots\\
        0& 0&\cdots&\psi_{i}(t_{n})
        \end{bmatrix}
    \end{split}
    \label{eq:psi_mtx}
\end{equation}
where \(t_{0},t_{1}, \cdots ,t_{n}\) indicate the corresponding times of demonstration trajectory points. Then, we can convert the terms $x(t)y_{offset}$ and $f_{des}(t)$ into a matrix form as $s$ and $F_d$:
\begin{equation}
    \begin{split}
        s = \begin{bmatrix}
            x(t_{0})y_{offset}\\
            \vdots\\
            x(t_{n})y_{offset}
        \end{bmatrix} ,\qquad        
        F_{d} = \begin{bmatrix}
            f_{des}(t_{0})\\
            \vdots\\
            f_{des}(t_{n})
        \end{bmatrix}   
    \end{split}
    \label{eq:sd_mtx}
\end{equation}

Substituting \autoref{eq:psi_mtx} and \autoref{eq:sd_mtx} into \autoref{eq:LWR_cost_t}, we can obtain:
\begin{equation}
    \begin{split}
        J(w_{i}) &= (F_{d} - s^Tw_{i})^T\Psi_{i}(F_{d} - s^Tw_{i})
    \end{split}
    \label{eq:LWR_mtx}
\end{equation}

Using the least square method to optimize, we obtain the learning weight along with time $t$ \cite{Schaal1998}:
\begin{equation}
    \begin{split}
        w_{i} = \frac{s^T\Psi_{i}F_{d}}{s^T\Psi_{i}s}
    \end{split}
    \label{eq:solve_w}
\end{equation}

\subsection{Learning From Demonstration Pipeline}
\label{sec:pipeline}

In our work, we grasp the needle using a script and utilize a simple trajectory planning technique to move the PSMs to the desired poses for handing over the needle. Therefore, we implement learning from demonstration algorithms for motion \romannum{2}\;and \romannum{4}\;as shown in section \ref{sec:3-C-2}.

\begin{figure}[tbh]
    \centering
    \includegraphics[width=0.9\linewidth]{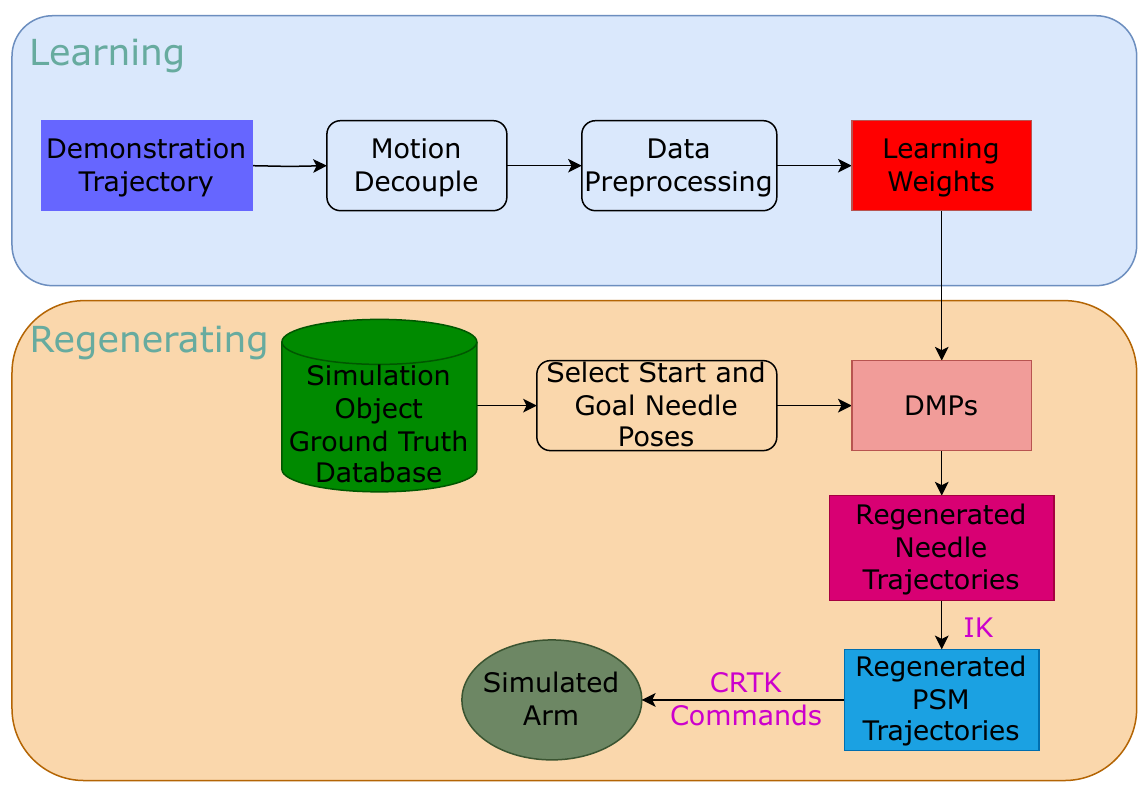}
    \caption{Learning from Demonstration pipeline}
    \label{fig:LfD_ppl}
\end{figure}

The pipeline for implementing the LfD algorithm is shown in \autoref{fig:LfD_ppl}.

The simulation object ground truth database contains the ground truth of all objects as described in section \ref{sec:3-A}. We can obtain the ground-truth states of the needle, entry \& exit points and PSM joints from both the demonstrations and the real-time simulation scene. Given those ground-truth states, we can find the desired start and goal states for the robot system's end-effector.

\section{Experiment Setup}

\subsection{Simulation Platform}
\label{sec:3-A}

\begin{figure}[tbh]
    \centering
    \subfloat[Phantom 1: Synthetic]
    {
    \includegraphics[width=0.45\linewidth, height=2.5cm]{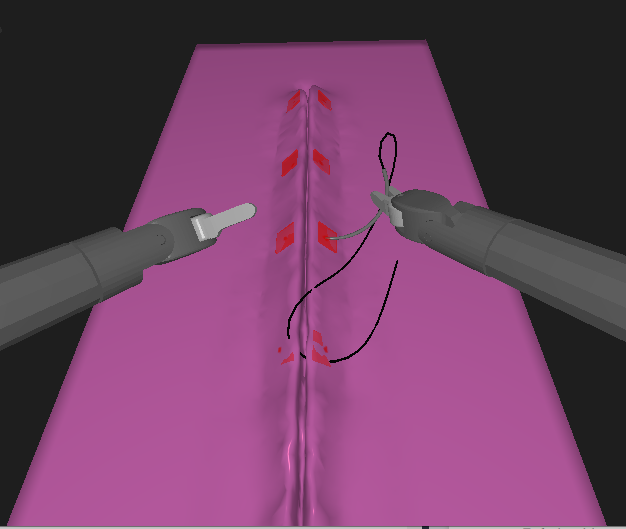}
    }
    \hfil
    \subfloat[Phantom 2: Scanned]
    {
    \includegraphics[width=0.45\linewidth, height=2.5cm]{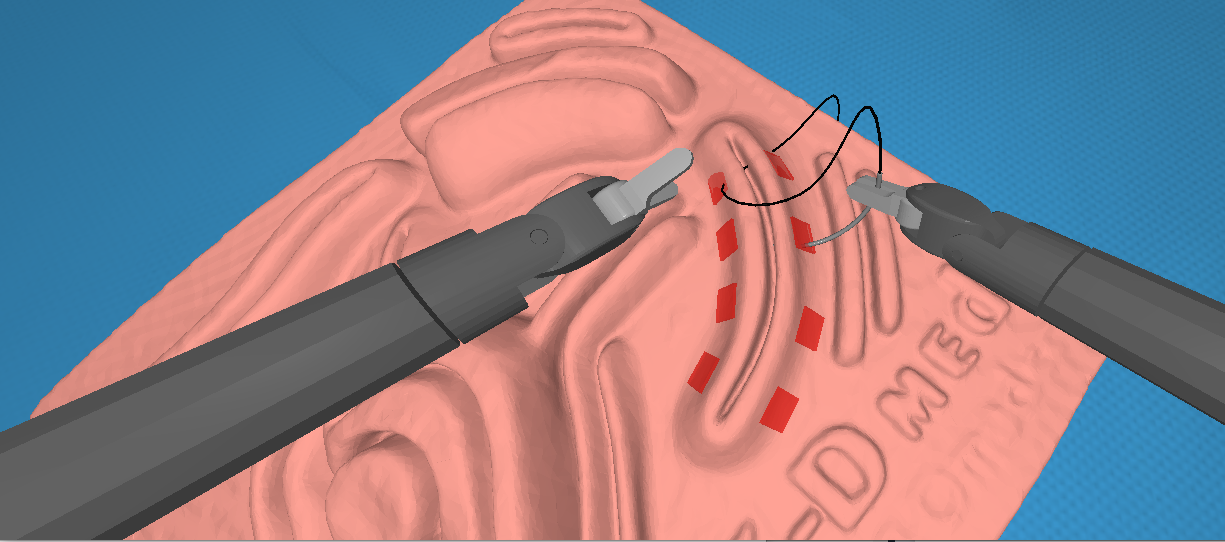}
    }
    \caption{Simulation environments}
    \label{fig:simulation_platforms}
\end{figure}

As shown in \autoref{fig:simulation_platforms}, we construct simulation scenes for suturing using AMBF\cite{munawar2022open}; the simulation platform contains:

\begin{itemize}
    \item a suturing phantom choosing from two alternatives
    \item two simulated dVRK Patient Side Manipulators (PSMs) with Large Needle Drivers from Intuitive Surgical, Inc.
    \item one simulated dVRK Endoscope Camera Manipulator (ECM) with a stereo camera attached
    \item one suture needle with radius of 10.18\,mm, 120-degree arc angle and thread attached
    \item red markers for entry and exit points
\end{itemize}

The dimensions of the dVRK arms are obtained from the measurements of the real-world first-generation dVSS.

\begin{figure}[tbh]
    \centering
    \subfloat[Real-World Phantom]
    {
    \includegraphics[width=0.45\linewidth, height=2.5cm]{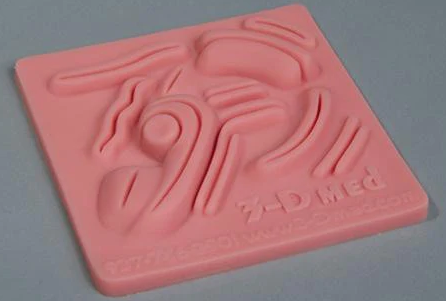}
    }
    \hfil
    \subfloat[MRI-Scanned Volume]
    {
    \includegraphics[width=0.45\linewidth, height=2.5cm]{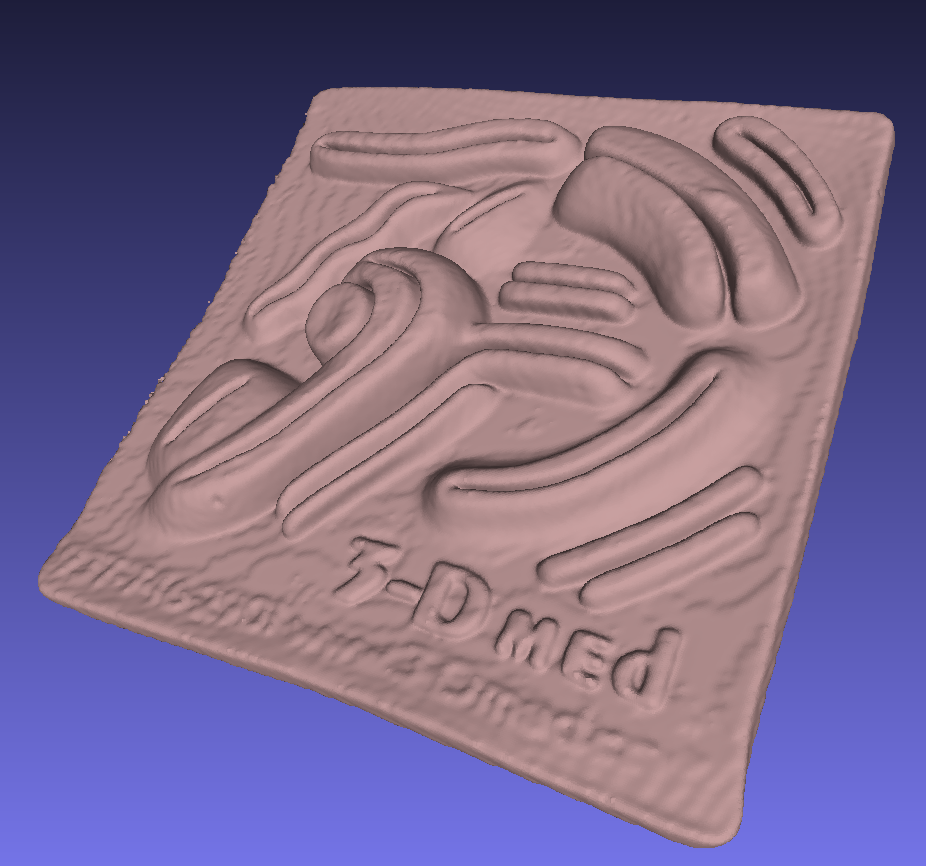}
    }
    \caption{3D-MED soft tissue suture pad}
    \label{fig:3dmed_phantom}
\end{figure}

To obtain phantom 2, we scanned the 3D-MED Soft Tissue Suture Pad using MRI and added the scanned volume to the simulation platform (\autoref{fig:3dmed_phantom}). Compared to CT scanning, MRI scanning can provide more details and has higher contrast for soft tissues or phantoms\cite{hiorns2011imaging}. Therefore, we selected MRI to scan the phantom and this method can be further used to scan real tissues or complex phantoms in the future.

\subsection{Teleoperation Setup}

\begin{figure}[tbh]
    \centering
    \subfloat[Physical dVRK]
    {
    \includegraphics[width=0.45\linewidth, height=3cm]{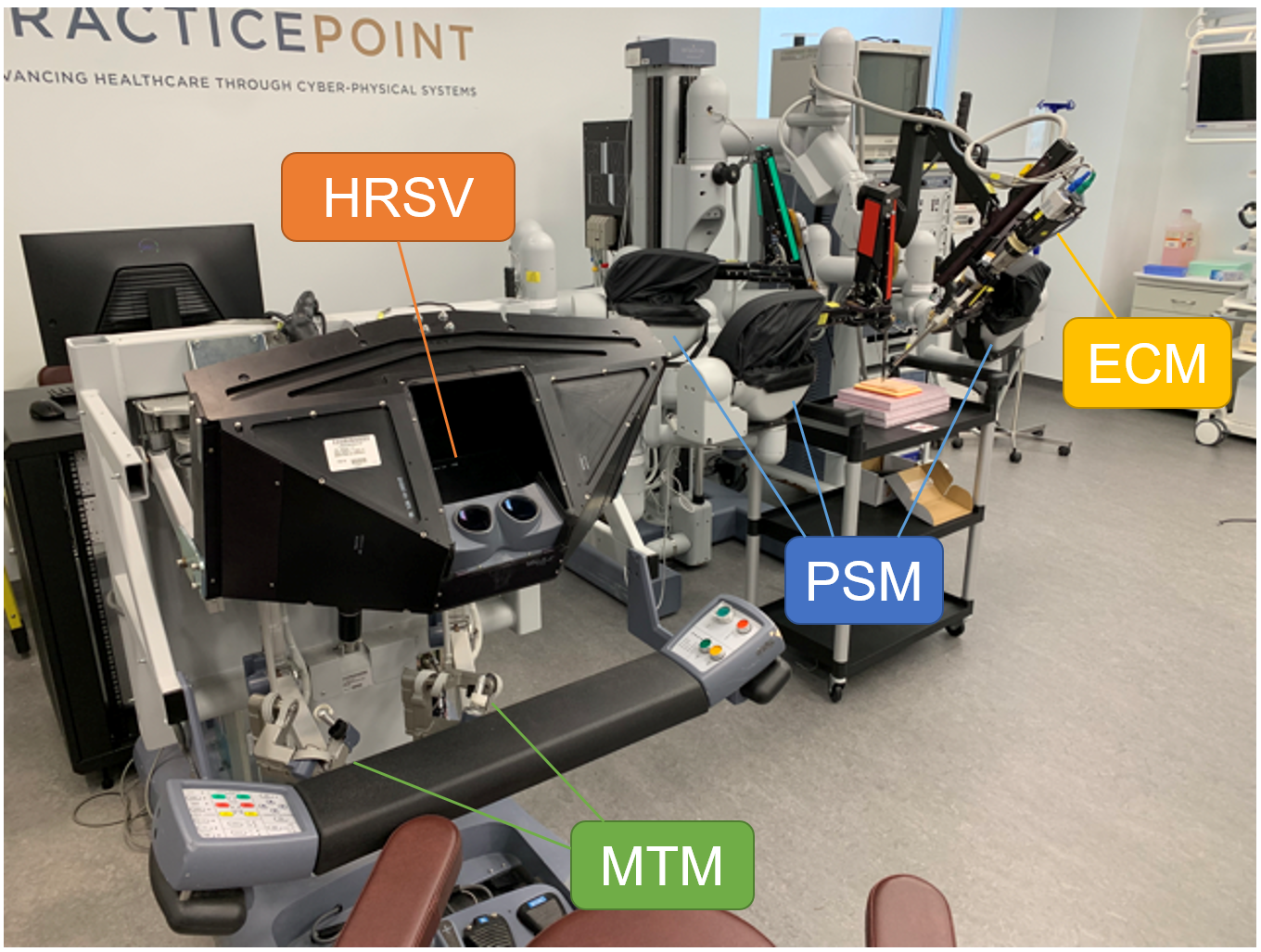}
    }
    \hfil
    \subfloat[Simulation Scene in HRSV]
    {
    \includegraphics[width=0.45\linewidth, height=3cm]{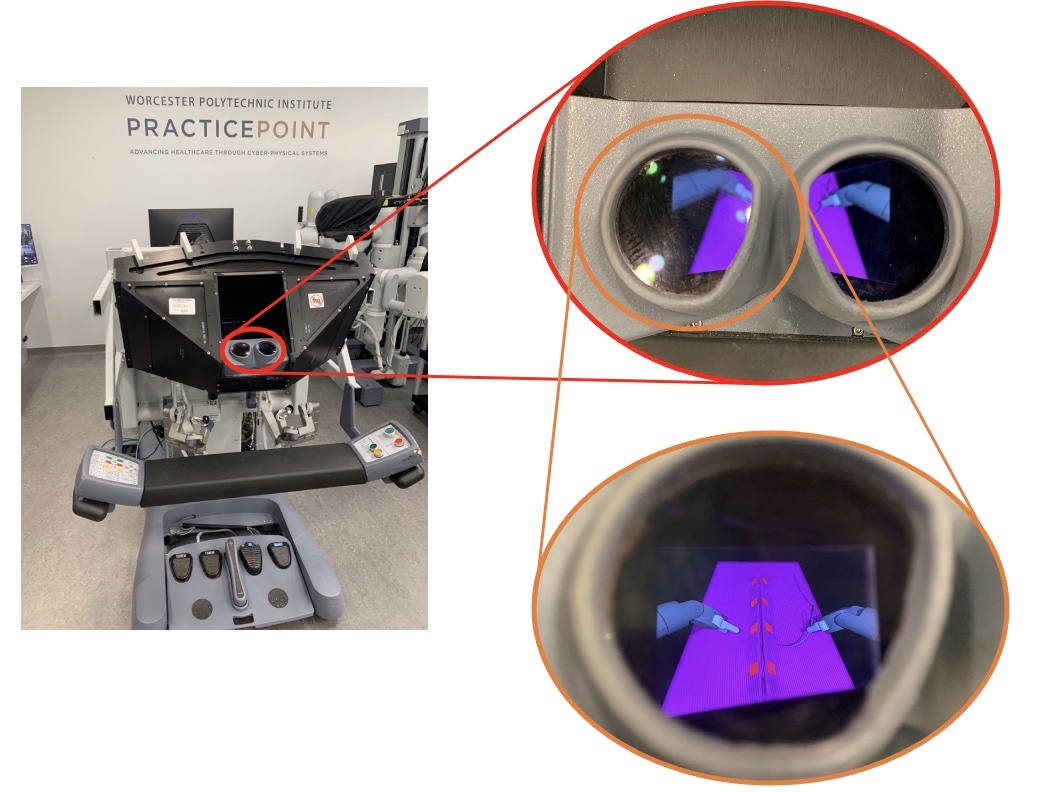}
    }
    \caption{Teleoperation setup}
    \label{fig:tele_setup}
\end{figure}

For the teleoperation setup, we utilize the dVRK High Resolution Stereo Viewer (HRSV, also known as the viewer console) and Master Tool Manipulators (MTMs), as shown in \autoref{fig:tele_setup}, to interact with the simulation platform. This setup is identical to the dVSS clinical setup and thus it can bring an immersive and realistic experience when teleoperating. 
Integrating with the dVRK HRSV brings the stereo vision to the system users. The stereo viewer allows the users to generate pseudo 3D vision so that they can have a more realistic and accurate sense of the depth information.

\subsection{User Study}
\label{sec:3-C-2}

For obtaining the training data for the LfD algorithm, we perform a user study under approved IRB protocols IRB-22-0593 at Worcester Polytechnic Institute and HIRB00000701 at Johns Hopkins University.  We recruited 10 subjects, consisting of 8 males and 2 females. Among those 10 users, 4 users have previous experience with surgical training and using the dVRK. The human subjects are asked to perform simple continuous sutures without tying knots in the simulation, as shown in the supplemental video. A single-throw suture without tying a knot can be decomposed into the following five subtasks\cite{schwaner2021autonomous}:
\begin{enumerate}[i.]
    \item Pick up the needle from the initial pose using the right arm (PSM2) and move PSM2 so that the needle tip is at the desired entry point
    \item Insert the needle through the phantom using PSM2 to the exit point
    \item Regrasp the needle with the left arm (PSM1).
    \item Extract the needle with PSM1, completing the throw
    \item Hand over the needle from PSM1 to PSM2 and move to the next initial pose
    \label{item: motion}
\end{enumerate}  

\begin{figure}[thb]
    \centering
    \subfloat[Phantom 1: Synthetic]{\includegraphics[width=0.45\linewidth, height=3.5cm]{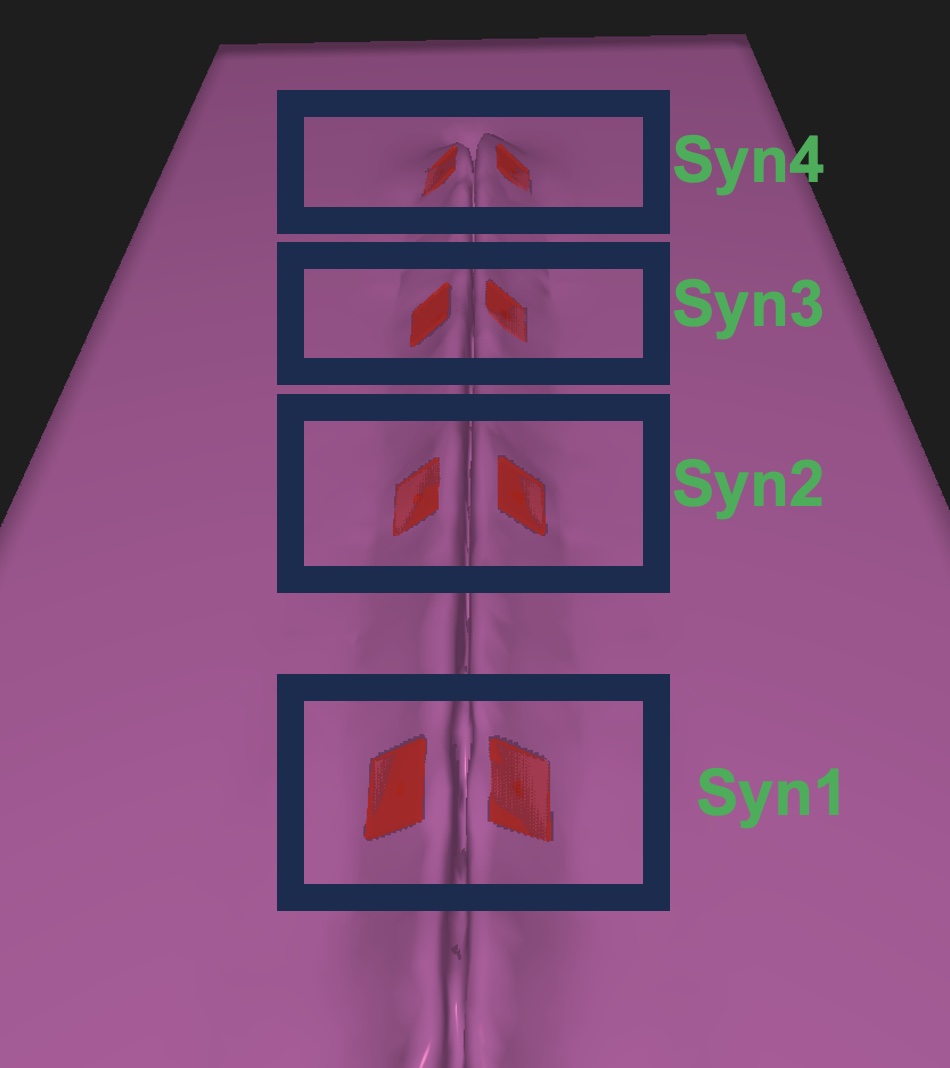}%
    \label{fig:6a}}
    \hfil
    \subfloat[Phantom 2: Scanned]{\includegraphics[width=0.45\linewidth,height=3.5cm]{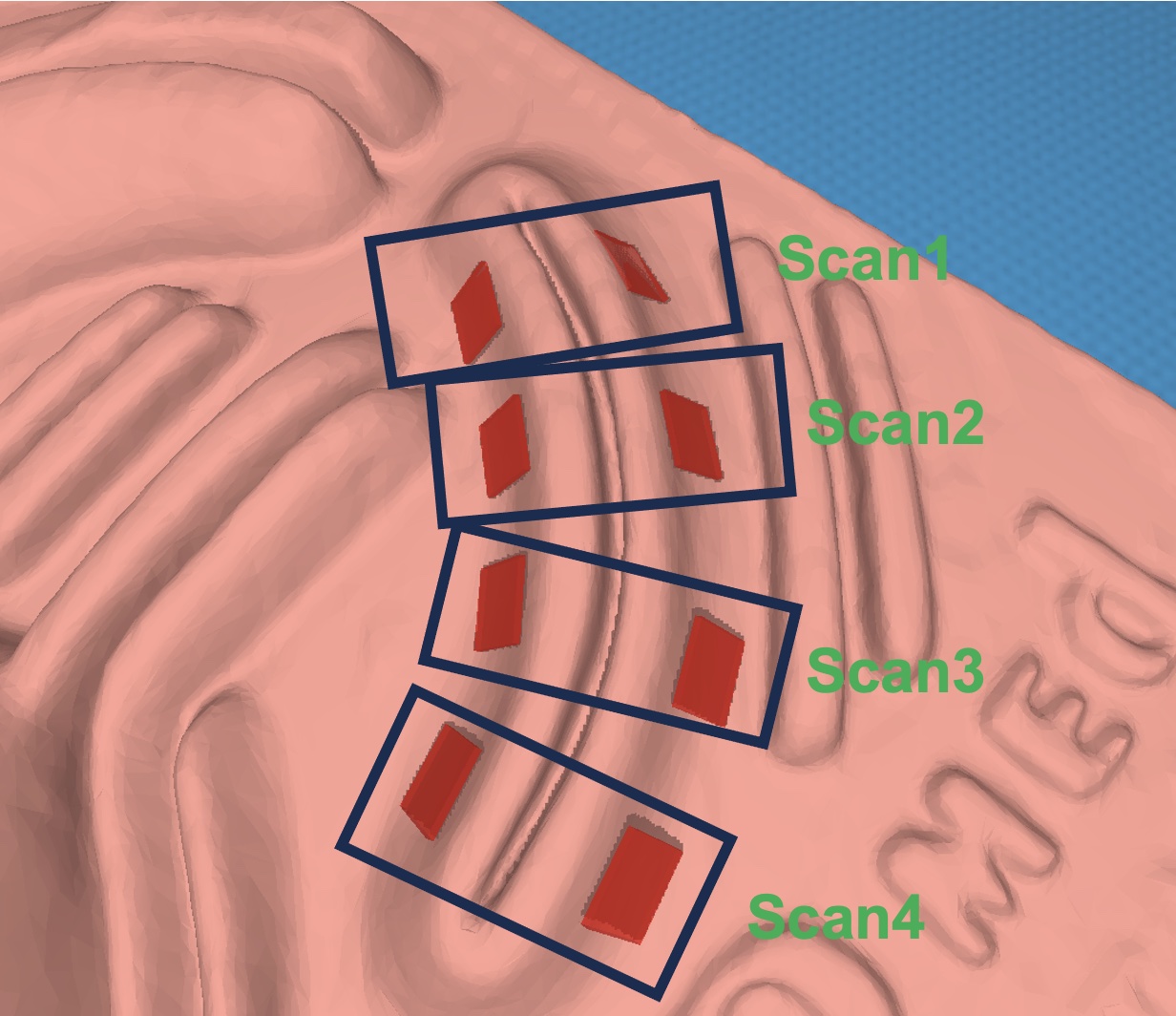}%
    \label{fig:6b}}
    \caption{Marker indices for phantoms}
    \label{fig:mark_idx}
\end{figure}

In the simulation scenes, we segmented the phantom and named different pairs of entry and exit markers as in \autoref{fig:mark_idx}.

The users perform the suture task following the sequence of indices from 1 to 4. For each phantom, the users complete the suture task for all 4 pairs consecutively. The collected data is segmented manually and utilized as the training datasets.

\subsection{Data Collection and Preprocessing}

In this section, we describe our data collection techniques and strategies.

\subsubsection{Data Collection Framework}

\begin{figure}[tbh]
    \centering
    \includegraphics[width=0.95\linewidth]{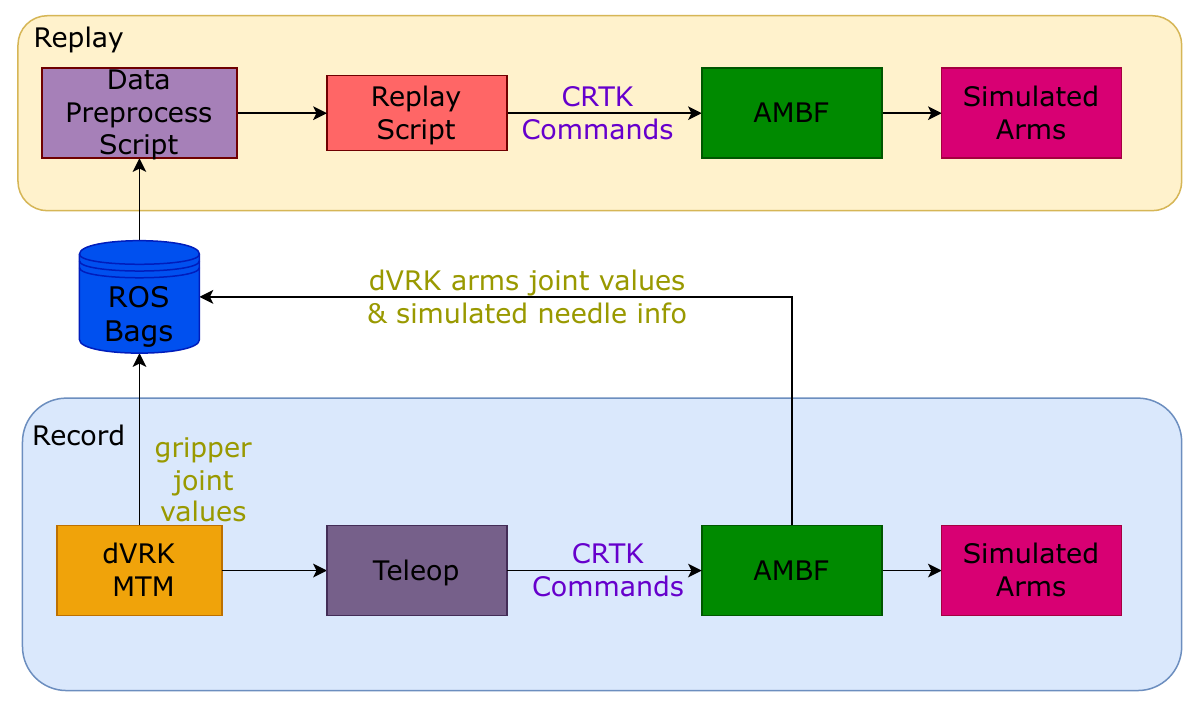}
    \caption{Motion data collection framework}
    \label{fig:data_frame}
\end{figure}

The control and communication commands of the simulation follow the Collaborative Robotics Toolkit (CRTK) convention \cite{su2020collaborative}, which ensures compatibility with the physical dVRK. To collect demonstration data, we develop the data collection framework shown in \autoref{fig:data_frame}.

The foot pedal presses are also included in the data collection. The raw collected data is in joint space and stored in ROS bags.

\subsubsection{Data Preprocessing}

The ground-truth poses of the needle, entry \& exit points are directly subscribed from the simulation. To segment the raw collected data, we use a velocity-based filter to examine the points with given thresholds \cite{iturrate2017learning}. When stretching out the raw input data, for Cartesian space positions, a quintic interpolation is used. On the other hand, for the Cartesian space orientations, spherical linear interpolation\cite{kremer2008quaternions} is used. For performance evaluation, we utilize the Dynamic Time Warping (DTW) technique \cite{kassidas1998synchronization, giorgino2009computing, vakanski2012trajectory} to synchronize trajectories with different lengths.

\section{Experimental Evaluation and Result}

\subsection{Model Parameters}
The hyperparameters we utilize in this paper are shown in \autoref{tab:model_para}.

\begin{table}[tbh]
    \centering
    \caption{Model hyper-parameters}
    \begin{tabular}{|c|c|c|}
        \hline
         Name & Value & Meaning \\
         \hline
         $\alpha_x$ & 1 & gain coefficient for the system variable $x$ \\
         \hline
         $\alpha_y$ / $\alpha_z$ & 25 & {proportional gain for position and orientation LfD} \\
         \hline
         $\beta_y$ / $\beta_z$ & 6.25 & {derivative gain for position and orientation LfD} \\
         \hline
         $N_{pts}$ & \makecell{500 \\ (100)} & \makecell{number of both LfD regenerated points for task \romannum{2} \\ (for task \romannum{4})}\\
         \hline
         $N_{bfs}$ & \makecell{100 \\ (50)} & \makecell{number of both LfD basis functions for task \romannum{2} \\ (for task \romannum{4})}\\
         \hline
         $N_{bfs}^o$ & \makecell{40 \\ (20) } & \makecell{number of both LfD basis functions for task \romannum{2} \\ (for task \romannum{4})}\\
         \hline
         
    \end{tabular}
    \label{tab:model_para}
\end{table}

\subsection{Trajectory Regeneration}
\begin{figure}[tbh]
    \centering
    \subfloat[Position Trajectory]{\includegraphics[width=0.45\linewidth, height=3.5cm]{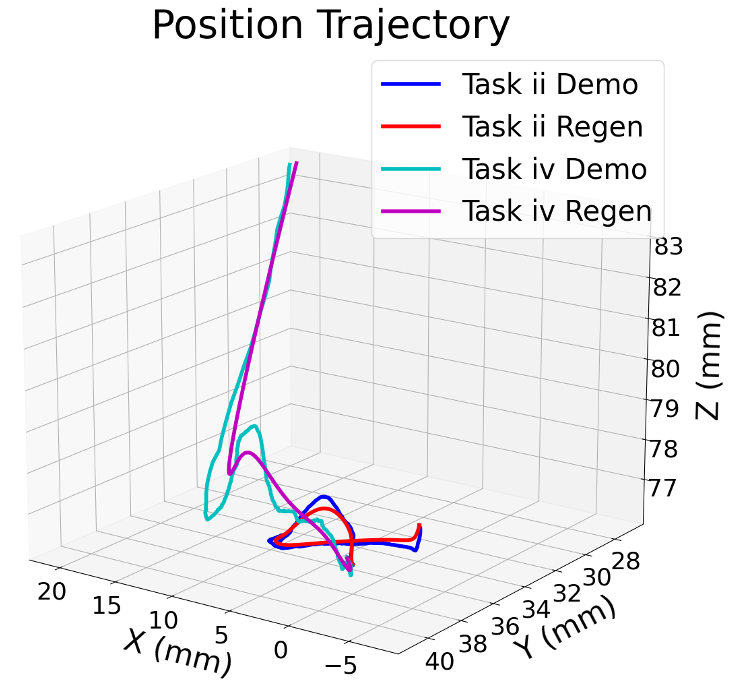}%
    \label{fig:7a}}
    \hfil
    \subfloat[Orientation Trajectory]{\includegraphics[width=0.45\linewidth, height=3.5cm]{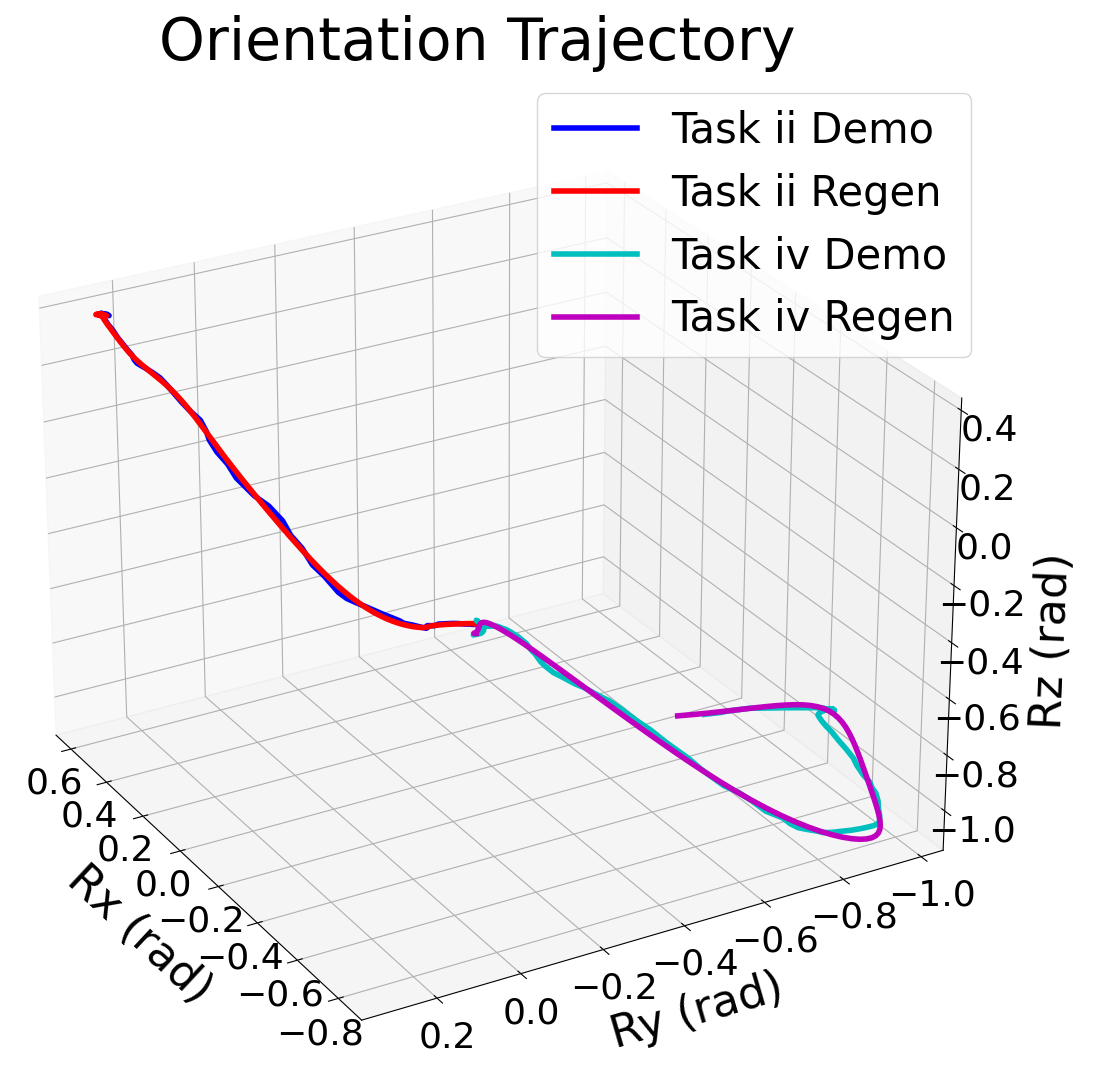}%
    \label{fig:7b}}
    \caption{Demonstration and regenerated trajectories of the suture needle at the pair \textit{\textbf{scan2}}}
    \label{fig:result_3d}
\end{figure}

First, we evaluate the performance of the LfD algorithm. For the evaluation, we select a random user's data at the \textit{\textbf{scan2}} pair of entry and exit markers. Then, we train the LfD algorithm and test on the same pair of entry and exit markers. After that, we plot both the demonstrations and the regenerated trajectories in \autoref{fig:result_3d}.
From this figure, we can anticipate that the regenerated trajectories can perform the same suture task in the simulation scenes successfully.

Furthermore, we repeat the above evaluation test for all 8 pairs of entry \& exit markers of all 10 users. We calculate not only the point-level errors like start and goal state errors but also the trajectory-level errors to assess the performance of the trajectory regeneration, as summarized in \autoref{tab:my_label}. Due to some abrupt shifts in the trajectory or among different users, we may have a chance to observe high average errors and can evaluate the median of the errors instead to access the performance.

\begin{table}[tbh]
    \centering
    \caption{Errors of LfD regenerated trajectories}
    \begin{tabular}{|c|c|c|c|}
        \hline
        \multicolumn{2}{|c|}{Item} & Mean & STD \\
        \hline
        \multirow{3}{*}{\makecell{task \romannum{2} \, pos error \\ (mm)}} & start & 0 & 0\\
        \cline{2-4}
        & goal & 0.05 & 0.06 \\
        \cline{2-4}
        & trajectory & 0.17 & 0.07 \\
        \hline
        \multirow{3}{*}{\makecell{task \romannum{2} \, ori error \\ (deg)}} & start & 0 & 0 \\
        \cline{2-4}
        & goal & 0.74 & 1.24 \\
        \cline{2-4}
        & trajectory & 1.71 & 8.44 \\
        \hline
        \multirow{3}{*}{\makecell{task \romannum{4} \, pos error \\ (mm)}} & start & 0.02 & 0.02 \\
        \cline{2-4}
        & goal & 0.43 & 0.31 \\
        \cline{2-4}
        & trajectory & 0.38 & 0.16 \\
        \hline
        \multirow{3}{*}{\makecell{task \romannum{4} \, ori error \\ (deg)}} & start & 0 & 0\\
        \cline{2-4}
        & goal & \makecell{11.32\\(median 2.71)} & 22.03 \\
        \cline{2-4}
        & trajectory & \makecell{17.54\\(median 1.42)} & 27.09 \\
        \hline
    \end{tabular}
    \label{tab:my_label}
\end{table}
When substituting the regenerated trajectories into the simulation scenes, 76 trajectories successfully accomplish the suture task. The trajectory regeneration using the LfD algorithm achieves 95\% success rate with reasonable start \& goal state errors.



\subsection{Generality Test}

For the next step, we assess the generality of the LfD algorithm by training the LfD algorithm on each pair of entry \& exit markers from both phantoms shown in \autoref{fig:mark_idx} and testing it using all pairs of entry \& exit markers. We observe the suture task completeness to assess the generality.

\begin{figure}[tbh]
    \centering
    \subfloat[Miss the exit marker]{\includegraphics[width=0.45\linewidth, height=2.5cm]{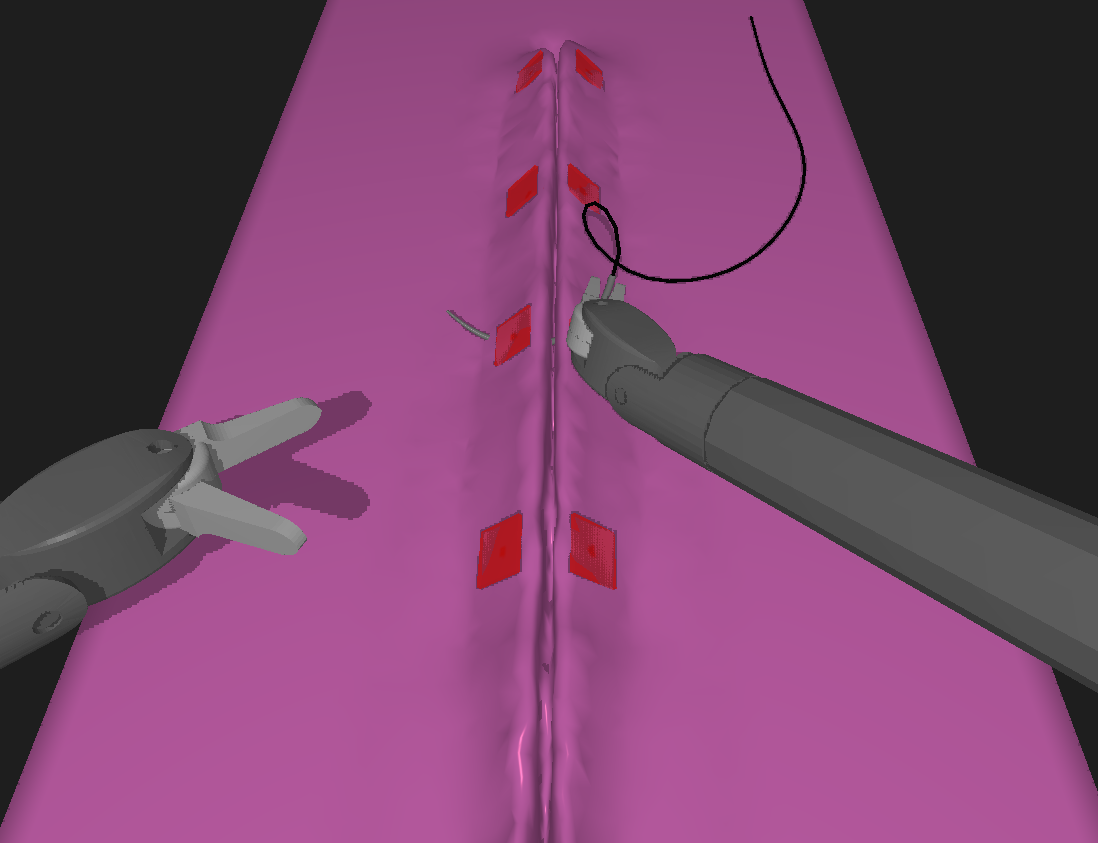}%
    \label{fig:8a}}
    \hfil
    \subfloat[Miss the entry marker]{\includegraphics[width=0.45\linewidth,height=2.5cm]{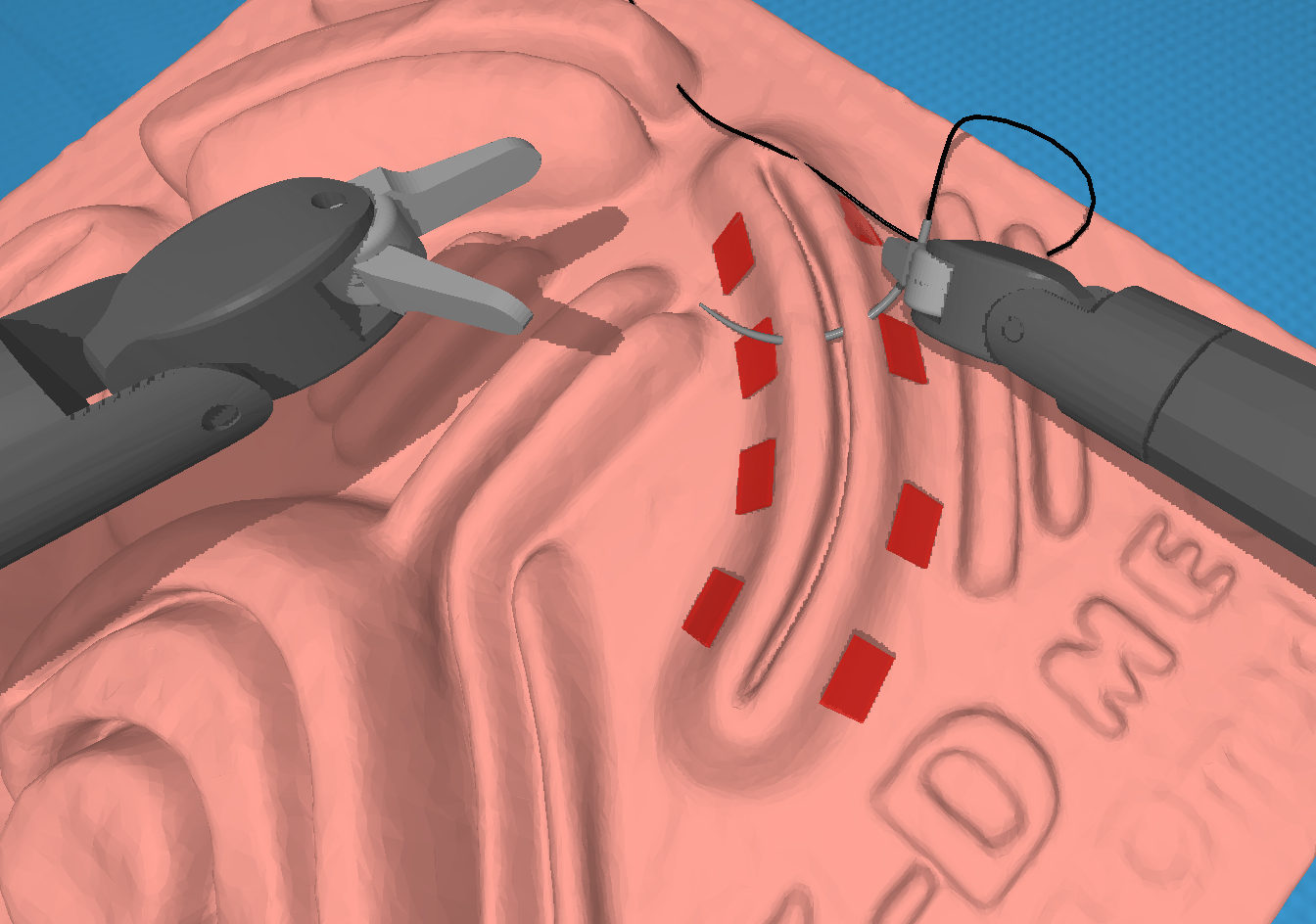}%
    \label{fig:8b}}
    \caption{Partial completion cases}
    \label{fig:part_success}
\end{figure}

The generality can be quantized based on the classical model of probability and represented in the following four levels:
\begin{itemize}
    \item 1.0 - Successfully complete the suture task
    \item 0.8 - Fail to complete the suture task, but only missing the exit marker when extracting, as shown in Figure\autoref{fig:8a}
    \item 0.4 - Fail to complete the suture task due to missing the entry marker when inserting but still following a reasonable trajectory, as shown in Figure\autoref{fig:8b}
    \item 0 - Fail to complete the suture task due to the other reasons
\end{itemize}

\begin{figure}[tbh]
    \centering
    \includegraphics[width=0.88\linewidth]{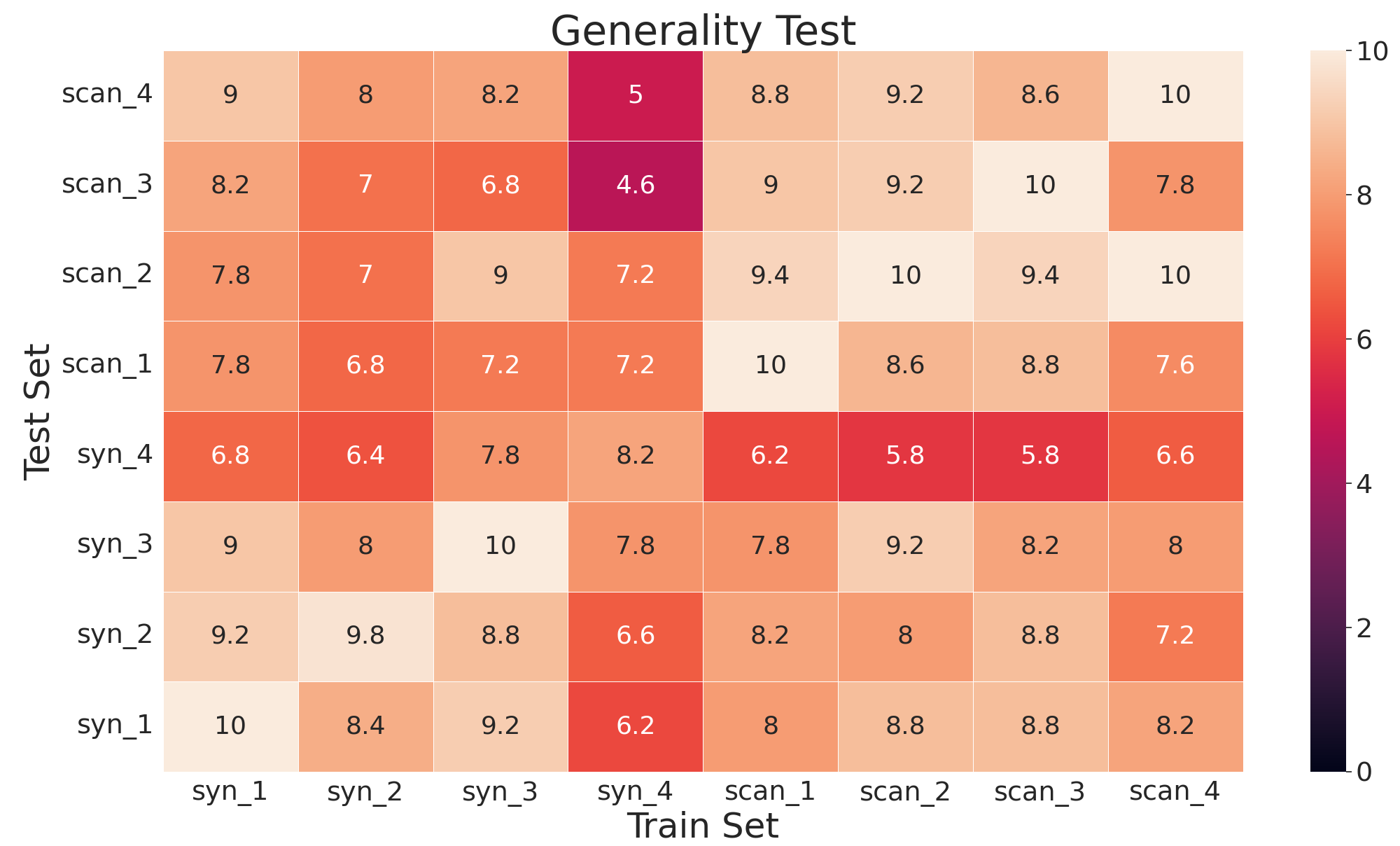}
    \caption{Generality test result}
    \label{fig:gen_all_heatmap}
\end{figure}

Going through the generality test for all users and all pairs of entry \& exit markers, we obtain the heat map shown in \autoref{fig:gen_all_heatmap}.

The average values of generality for all users are shown in \autoref{tab:general_result}. We can see that the LfD algorithm has reasonable generality for the selected suturing subtasks and is worth further investigation. In addition, we can anticipate that we may have a chance to achieve better performance by introducing more experienced or even professional human subjects.

\begin{table}[tbh]
\begin{minipage}[t]{0.45\linewidth}
    \caption{Generality result}
    \centering
    \begin{tabular}{|c|c|}
        \hline
         \textbf{User} & \textbf{Generality}  \\
         \hline
         Experienced & 0.915 \\
         \hline
         Naive & 0.742 \\
         \hline
         Overall & 0.811\\
         \hline
    \end{tabular}
    \label{tab:general_result}
\end{minipage}
\hfill
\begin{minipage}[t]{0.54\linewidth}
    \caption{Success rate of experienced users}
    \centering
    \begin{tabular}{|c|c|c|}
        \hline
         \textbf{Task} & \textbf{Individual} & \textbf{Overall} \\
         \hline
         \romannum{1} & 256\,/\,256 & 256\,/\,256 \\
         \hline
         \romannum{2} & 202\,/\,256 & 202\,/\,256\\
         \hline
         \romannum{3} & 197\,/\,202 & 197\,/\,256 \\
         \hline
         \romannum{4} & 181\,/\,197 & 181\,/\,256 \\
         \hline
         \romannum{5} & 181\,/\,181 & 181\,/\,256 \\
         \hline
    \end{tabular}
    \label{tab:exp_success_rate}
\end{minipage}
\end{table}

\subsection{Task Execution Performance}

According to the results shown in \autoref{tab:general_result}, we can see that the regenerated trajectories learned from the experienced users have much better performance in the generality test. Therefore, when assessing the overall suturing task success rate, we will only focus on the experienced users set to exclude the errors due to lack of acquaintance of the skills.

From \autoref{tab:exp_success_rate}, we can see that the regenerated trajectories learned from the experienced users have an overall success rate of 70.7\% for suturing task completion. Nevertheless, we also find that most of the failures have a similar scenario as shown in Figure\autoref{fig:8a}. For success rate evaluation, the result is binary instead of discrete values. Therefore, compared to the generality test result, the success rates would have smaller values due to the false negative cases.

Last but not least, we evaluate the task execution time. We record the time taken for completing a whole suturing procedure on each pair of entry \& exit markers and obtain:
\begin{figure}[H]
    \centering
    \includegraphics[width=0.7\linewidth]{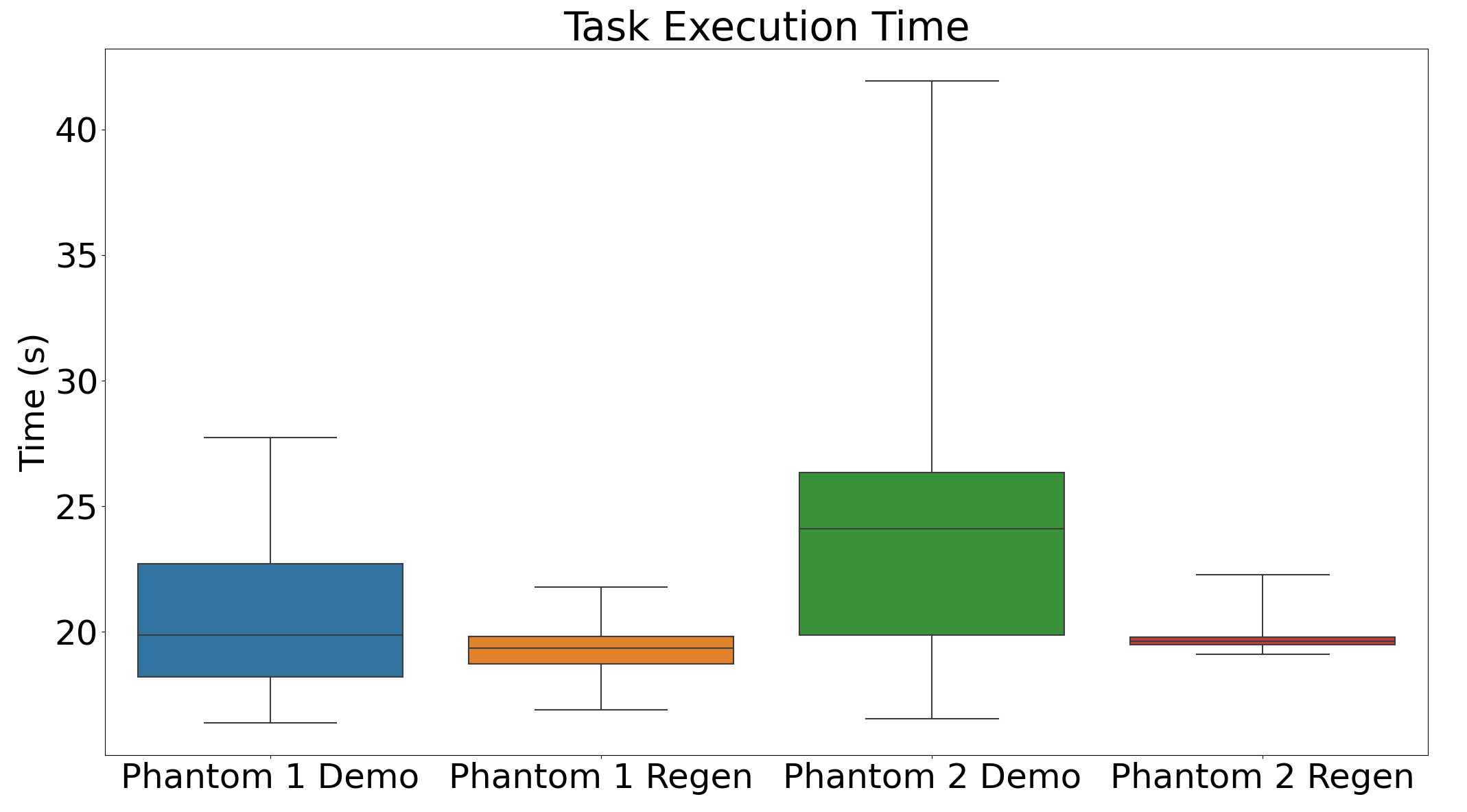}
    \caption{Task execution time}
    \label{fig:task_time}
\end{figure}
From \autoref{fig:task_time}, we can see that for suturing tasks on a simple synthetic phantom, the LfD algorithm can slightly improve the time efficiency when completing the tasks. On the other hand, for the tasks on the scanned phantom with higher complexity, the LfD algorithm can reduce the total time for completing the suturing task by 20\%. 

\section{Discussion and Future Development}

In this work, we build a novel and realistic simulation scene using an MRI-scanned phantom, and construct a data collection pipeline for the simulation. Also, we present a LfD algorithm using DMP and LWR for suturing task automation with comprehensive analyses. As a result, we can see that the regenerated trajectories using the LfD algorithm can complete the suturing task with 95\% success rate in the simulation. Also, the algorithm achieves high generality of 0.811 and time efficiency by a 20\% reduction in task execution time. In addition, we can see that the regenerated trajectories learned from the experienced users usually have better performance. Therefore, for further development, we can introduce more experienced human subjects or professional human subjects such as surgeons. Moreover, we can implement skill assessment techniques \cite{moorthy2004bimodal, frischknecht2013objective, vedula2016analysis, ijgosse2020competency, shayan2023measuring} when preprocessing the data to exclude unskilled demonstrations.

Taking advantage of AMBF and CRTK, and integrating with advanced vision perception methods \cite{zhong2016adaptive, chiu2021bimanual, chiu2022markerless, jiang2023development, gao2023intraoperative, qian2020flexivision, gao2023laparoscopic} to obtain the estimated poses of the suturing needle, we may further extend the proposed algorithm from the simulation to the real world.

\section*{Acknowledgement}

This work was supported in part by NSF AccelNet awards OISE-1927275 and OISE-1927354. Thanks to Juan Antonio Barragan, Hisashi Ishida and Adnan Munawar for their contributions toward establishing the simulation framework.





\bibliographystyle{IEEEtran}
\bibliography{references}

\begin{thebibliography}{10}
\providecommand{\url}[1]{#1}
\csname url@samestyle\endcsname
\providecommand{\newblock}{\relax}
\providecommand{\bibinfo}[2]{#2}
\providecommand{\BIBentrySTDinterwordspacing}{\spaceskip=0pt\relax}
\providecommand{\BIBentryALTinterwordstretchfactor}{4}
\providecommand{\BIBentryALTinterwordspacing}{\spaceskip=\fontdimen2\font plus
\BIBentryALTinterwordstretchfactor\fontdimen3\font minus
  \fontdimen4\font\relax}
\providecommand{\BIBforeignlanguage}[2]{{%
\expandafter\ifx\csname l@#1\endcsname\relax
\typeout{** WARNING: IEEEtran.bst: No hyphenation pattern has been}%
\typeout{** loaded for the language `#1'. Using the pattern for}%
\typeout{** the default language instead.}%
\else
\language=\csname l@#1\endcsname
\fi
#2}}
\providecommand{\BIBdecl}{\relax}
\BIBdecl

\bibitem{sheetz2020trends}
K.~H. Sheetz, J.~Claflin, and J.~B. Dimick, ``Trends in the adoption of robotic
  surgery for common surgical procedures,'' \emph{JAMA network open}, vol.~3,
  no.~1, pp. e1\,918\,911--e1\,918\,911, 2020.

\bibitem{attanasio2021autonomy}
A.~Attanasio, B.~Scaglioni, E.~De~Momi, P.~Fiorini, and P.~Valdastri,
  ``Autonomy in surgical robotics,'' \emph{Annual Review of Control, Robotics,
  and Autonomous Systems}, vol.~4, pp. 651--679, 2021.

\bibitem{d2004robotic}
A.~D’Annibale, E.~Morpurgo, V.~Fiscon, P.~Trevisan, G.~Sovernigo, C.~Orsini,
  and D.~Guidolin, ``Robotic and laparoscopic surgery for treatment of
  colorectal diseases,'' \emph{Diseases of the colon \& rectum}, vol.~47, pp.
  2162--2168, 2004.

\bibitem{kazanzides2014open}
P.~Kazanzides, Z.~Chen, A.~Deguet, G.~S. Fischer, R.~H. Taylor, and S.~P.
  DiMaio, ``An open-source research kit for the da vinci{\textregistered}
  surgical system,'' in \emph{2014 IEEE international conference on robotics
  and automation (ICRA)}.\hskip 1em plus 0.5em minus 0.4em\relax IEEE, 2014,
  pp. 6434--6439.

\bibitem{munawar2019real}
A.~Munawar, Y.~Wang, R.~Gondokaryono, and G.~S. Fischer, ``A real-time dynamic
  simulator and an associated front-end representation format for simulating
  complex robots and environments,'' in \emph{2019 IEEE/RSJ International
  Conference on Intelligent Robots and Systems (IROS)}.\hskip 1em plus 0.5em
  minus 0.4em\relax IEEE, 2019, pp. 1875--1882.

\bibitem{munawar2022open}
A.~Munawar, J.~Y. Wu, G.~S. Fischer, R.~H. Taylor, and P.~Kazanzides, ``Open
  simulation environment for learning and practice of robot-assisted surgical
  suturing,'' \emph{IEEE Robotics and Automation Letters}, vol.~7, no.~2, pp.
  3843--3850, 2022.

\bibitem{Ijspeert2002}
A.~J. Ijspeert, J.~Nakanishi, and S.~Schaal, ``Movement imitation with
  nonlinear dynamical systems in humanoid robots,'' in \emph{Proceedings 2002
  IEEE International Conference on Robotics and Automation (Cat. No.
  02CH37292)}, vol.~2.\hskip 1em plus 0.5em minus 0.4em\relax IEEE, 2002, pp.
  1398--1403.

\bibitem{Schaal2006}
S.~Schaal, ``Dynamic movement primitives-a framework for motor control in
  humans and humanoid robotics,'' in \emph{Adaptive motion of animals and
  machines}.\hskip 1em plus 0.5em minus 0.4em\relax Springer, 2006, pp.
  261--280.

\bibitem{ijspeert2013dynamical}
A.~J. Ijspeert, J.~Nakanishi, H.~Hoffmann, P.~Pastor, and S.~Schaal,
  ``Dynamical movement primitives: learning attractor models for motor
  behaviors,'' \emph{Neural computation}, vol.~25, no.~2, pp. 328--373, 2013.

\bibitem{Schaal1998}
S.~Schaal, C.~G. Atkeson, and S.~Vijayakumar, ``Scalable techniques from
  nonparametric statistics for real time robot learning,'' \emph{Applied
  Intelligence}, vol.~17, no.~1, pp. 49--60, 2002.

\bibitem{sen2016automating}
S.~Sen, A.~Garg, D.~V. Gealy, S.~McKinley, Y.~Jen, and K.~Goldberg,
  ``Automating multi-throw multilateral surgical suturing with a mechanical
  needle guide and sequential convex optimization,'' in \emph{2016 IEEE
  international conference on robotics and automation (ICRA)}.\hskip 1em plus
  0.5em minus 0.4em\relax IEEE, 2016, pp. 4178--4185.

\bibitem{varier2020collaborative}
V.~M. Varier, D.~K. Rajamani, N.~Goldfarb, F.~Tavakkolmoghaddam, A.~Munawar,
  and G.~S. Fischer, ``Collaborative suturing: A reinforcement learning
  approach to automate hand-off task in suturing for surgical robots,'' in
  \emph{2020 29th IEEE international conference on robot and human interactive
  communication (RO-MAN)}.\hskip 1em plus 0.5em minus 0.4em\relax IEEE, 2020,
  pp. 1380--1386.

\bibitem{schwaner2021autonomous_needle}
K.~L. Schwaner, D.~Dall'Alba, P.~T. Jensen, P.~Fiorini, and T.~R. Savarimuthu,
  ``Autonomous needle manipulation for robotic surgical suturing based on
  skills learned from demonstration,'' in \emph{2021 IEEE 17th international
  conference on automation science and engineering (CASE)}.\hskip 1em plus
  0.5em minus 0.4em\relax IEEE, 2021, pp. 235--241.

\bibitem{schwaner2021autonomous}
K.~L. Schwaner, I.~Iturrate, J.~K. Andersen, P.~T. Jensen, and T.~R.
  Savarimuthu, ``Autonomous bi-manual surgical suturing based on skills learned
  from demonstration,'' in \emph{2021 IEEE/RSJ International Conference on
  Intelligent Robots and Systems (IROS)}.\hskip 1em plus 0.5em minus
  0.4em\relax IEEE, 2021, pp. 4017--4024.

\bibitem{allard2007sofa}
J.~Allard, S.~Cotin, F.~Faure, P.-J. Bensoussan, F.~Poyer, C.~Duriez,
  H.~Delingette, and L.~Grisoni, ``Sofa-an open source framework for medical
  simulation,'' in \emph{MMVR 15-Medicine Meets Virtual Reality}, vol.
  125.\hskip 1em plus 0.5em minus 0.4em\relax IOP Press, 2007, pp. 13--18.

\bibitem{lungu2021review}
A.~J. Lungu, W.~Swinkels, L.~Claesen, P.~Tu, J.~Egger, and X.~Chen, ``A review
  on the applications of virtual reality, augmented reality and mixed reality
  in surgical simulation: an extension to different kinds of surgery,''
  \emph{Expert review of medical devices}, vol.~18, no.~1, pp. 47--62, 2021.

\bibitem{haiderbhai2022robust}
M.~Haiderbhai, R.~Gondokaryono, T.~Looi, J.~M. Drake, and L.~A. Kahrs, ``Robust
  sim2real transfer with the da vinci research kit: A study on camera,
  lighting, and physics domain randomization,'' in \emph{2022 IEEE/RSJ
  International Conference on Intelligent Robots and Systems (IROS)}.\hskip 1em
  plus 0.5em minus 0.4em\relax IEEE, 2022, pp. 3429--3435.

\bibitem{jiang2023markerless}
Y.~Jiang, H.~Zhou, and G.~S. Fischer, ``Markerless suture needle tracking from
  a robotic endoscope based on deep learning,'' in \emph{2023 International
  Symposium on Medical Robotics (ISMR)}.\hskip 1em plus 0.5em minus 0.4em\relax
  IEEE, 2023, pp. 1--7.

\bibitem{ijspeert2003learning}
A.~J. Ijspeert, J.~Nakanishi, and S.~Schaal, ``Learning attractor landscapes
  for learning motor primitives,'' in \emph{Advances in neural information
  processing systems}, 2003, pp. 1547--1554.

\bibitem{schaal2003control}
S.~Schaal, J.~Peters, J.~Nakanishi, and A.~Ijspeert, ``Control, planning,
  learning, and imitation with dynamic movement primitives,'' in \emph{Workshop
  on Bilateral Paradigms on Humans and Humanoids: IEEE International Conference
  on Intelligent Robots and Systems (IROS 2003)}, 2003, pp. 1--21.

\bibitem{Pervez2018}
A.~Pervez and D.~Lee, ``Learning task-parameterized dynamic movement primitives
  using mixture of gmms,'' \emph{Intelligent Service Robotics}, vol.~11, no.~1,
  pp. 61--78, 2018.

\bibitem{ginesi2020autonomous}
M.~Ginesi, D.~Meli, A.~Roberti, N.~Sansonetto, and P.~Fiorini, ``Autonomous
  task planning and situation awareness in robotic surgery,'' in \emph{2020
  IEEE/RSJ International Conference on Intelligent Robots and Systems
  (IROS)}.\hskip 1em plus 0.5em minus 0.4em\relax IEEE, 2020, pp. 3144--3150.

\bibitem{chiaverini1999unit}
S.~Chiaverini, B.~Siciliano \emph{et~al.}, ``Unit quaternion: a useful tool for
  inverse kinematics of robot manipulators,'' \emph{SYSTEMS ANALYSIS,
  MODELLING, SIMULATION}, vol.~35, no.~1, pp. 45--60, 1999.

\bibitem{ude1999filtering}
A.~Ude, ``Filtering in a unit quaternion space for model-based object
  tracking,'' \emph{Robotics and Autonomous Systems}, vol.~28, no. 2-3, pp.
  163--172, 1999.

\bibitem{sabatini2006quaternion}
A.~M. Sabatini, ``Quaternion-based extended kalman filter for determining
  orientation by inertial and magnetic sensing,'' \emph{IEEE transactions on
  Biomedical Engineering}, vol.~53, no.~7, pp. 1346--1356, 2006.

\bibitem{faraway2009modelling}
J.~J. Faraway and S.~B. Choe, ``Modelling orientation trajectories,''
  \emph{Statistical Modelling}, vol.~9, no.~1, pp. 51--68, 2009.

\bibitem{Gams}
A.~Ude, B.~Nemec, T.~Petri{\'c}, and J.~Morimoto, ``Orientation in cartesian
  space dynamic movement primitives,'' in \emph{2014 IEEE International
  Conference on Robotics and Automation (ICRA)}.\hskip 1em plus 0.5em minus
  0.4em\relax IEEE, 2014, pp. 2997--3004.

\bibitem{hiorns2011imaging}
M.~P. Hiorns, ``Imaging of the urinary tract: the role of ct and mri,''
  \emph{Pediatric nephrology}, vol.~26, no.~1, pp. 59--68, 2011.

\bibitem{su2020collaborative}
Y.-H. Su, A.~Munawar, A.~Deguet, A.~Lewis, K.~Lindgren, Y.~Li, R.~H. Taylor,
  G.~S. Fischer, B.~Hannaford, and P.~Kazanzides, ``Collaborative robotics
  toolkit (crtk): Open software framework for surgical robotics research,'' in
  \emph{2020 Fourth IEEE International Conference on Robotic Computing
  (IRC)}.\hskip 1em plus 0.5em minus 0.4em\relax IEEE, 2020, pp. 48--55.

\bibitem{iturrate2017learning}
I.~Iturrate, E.~H. {\O}stergaard, M.~Rytter, and T.~R. Savarimuthu, ``Learning
  and correcting robot trajectory keypoints from a single demonstration,'' in
  \emph{2017 3rd International Conference on Control, Automation and Robotics
  (ICCAR)}.\hskip 1em plus 0.5em minus 0.4em\relax IEEE, 2017, pp. 52--59.

\bibitem{kremer2008quaternions}
V.~E. Kremer, ``Quaternions and slerp,'' in \emph{Embots. dfki. de/doc/seminar
  ca/Kremer Quaternions. pdf}, 2008.

\bibitem{kassidas1998synchronization}
A.~Kassidas, J.~F. MacGregor, and P.~A. Taylor, ``Synchronization of batch
  trajectories using dynamic time warping,'' \emph{AIChE Journal}, vol.~44,
  no.~4, pp. 864--875, 1998.

\bibitem{giorgino2009computing}
T.~Giorgino, ``Computing and visualizing dynamic time warping alignments in r:
  the dtw package,'' \emph{Journal of statistical Software}, vol.~31, pp.
  1--24, 2009.

\bibitem{vakanski2012trajectory}
A.~Vakanski, I.~Mantegh, A.~Irish, and F.~Janabi-Sharifi, ``Trajectory learning
  for robot programming by demonstration using hidden markov model and dynamic
  time warping,'' \emph{IEEE Transactions on Systems, Man, and Cybernetics,
  Part B (Cybernetics)}, vol.~42, no.~4, pp. 1039--1052, 2012.

\bibitem{moorthy2004bimodal}
K.~Moorthy, Y.~Munz, A.~Dosis, F.~Bello, A.~Chang, and A.~Darzi, ``Bimodal
  assessment of laparoscopic suturing skills: construct and concurrent
  validity,'' \emph{Surgical Endoscopy And Other Interventional Techniques},
  vol.~18, pp. 1608--1612, 2004.

\bibitem{frischknecht2013objective}
A.~C. Frischknecht, S.~J. Kasten, S.~J. Hamstra, N.~C. Perkins, R.~B.
  Gillespie, T.~J. Armstrong, and R.~M. Minter, ``The objective assessment of
  experts’ and novices’ suturing skills using an image analysis program,''
  \emph{Academic Medicine}, vol.~88, no.~2, pp. 260--264, 2013.

\bibitem{vedula2016analysis}
S.~S. Vedula, A.~O. Malpani, L.~Tao, G.~Chen, Y.~Gao, P.~Poddar, N.~Ahmidi,
  C.~Paxton, R.~Vidal, S.~Khudanpur \emph{et~al.}, ``Analysis of the structure
  of surgical activity for a suturing and knot-tying task,'' \emph{PloS one},
  vol.~11, no.~3, p. e0149174, 2016.

\bibitem{ijgosse2020competency}
W.~M. IJgosse, E.~Leijte, S.~Ganni, J.-M. Luursema, N.~K. Francis, J.~J.
  Jakimowicz, and S.~M. Botden, ``Competency assessment tool for laparoscopic
  suturing: development and reliability evaluation,'' \emph{Surgical
  Endoscopy}, vol.~34, pp. 2947--2953, 2020.

\bibitem{shayan2023measuring}
A.~M. Shayan, S.~Singh, J.~Gao, R.~E. Groff, J.~Bible, J.~F. Eidt, M.~Sheahan,
  S.~S. Gandhi, J.~V. Blas, and R.~Singapogu, ``Measuring hand movement for
  suturing skill assessment: A simulation-based study,'' \emph{Surgery}, vol.
  174, no.~5, pp. 1184--1192, 2023.

\bibitem{zhong2016adaptive}
F.~Zhong, D.~Navarro-Alarcon, Z.~Wang, Y.-h. Liu, T.~Zhang, H.~M. Yip, and
  H.~Wang, ``Adaptive 3d pose computation of suturing needle using constraints
  from static monocular image feedback,'' in \emph{2016 IEEE/RSJ International
  Conference on Intelligent Robots and Systems (IROS)}.\hskip 1em plus 0.5em
  minus 0.4em\relax IEEE, 2016, pp. 5521--5526.

\bibitem{chiu2021bimanual}
Z.-Y. Chiu, F.~Richter, E.~K. Funk, R.~K. Orosco, and M.~C. Yip, ``Bimanual
  regrasping for suture needles using reinforcement learning for rapid motion
  planning,'' in \emph{2021 IEEE International Conference on Robotics and
  Automation (ICRA)}.\hskip 1em plus 0.5em minus 0.4em\relax IEEE, 2021, pp.
  7737--7743.

\bibitem{chiu2022markerless}
Z.-Y. Chiu, A.~Z. Liao, F.~Richter, B.~Johnson, and M.~C. Yip, ``Markerless
  suture needle 6d pose tracking with robust uncertainty estimation for
  autonomous minimally invasive robotic surgery,'' in \emph{2022 IEEE/RSJ
  International Conference on Intelligent Robots and Systems (IROS)}.\hskip 1em
  plus 0.5em minus 0.4em\relax IEEE, 2022, pp. 5286--5292.

\bibitem{jiang2023development}
Y.~Jiang, H.~Zhou, and G.~S. Fischer, ``Development and evaluation of a
  markerless 6 dof pose tracking method for a suture needle from a robotic
  endoscope,'' \emph{Journal of Medical Robotics Research}, vol.~8, no. 03n04,
  p. 2340009, 2023.

\bibitem{gao2023intraoperative}
S.~Gao, Y.~Wang, X.~Ma, H.~Zhou, Y.~Jiang, K.~Yang, L.~Lu, S.~Wang, B.~C.
  Nephew, L.~Fichera \emph{et~al.}, ``Intraoperative laparoscopic photoacoustic
  image guidance system in the da vinci surgical system,'' \emph{Biomedical
  optics express}, vol.~14, no.~9, pp. 4914--4928, 2023.

\bibitem{qian2020flexivision}
L.~Qian, C.~Song, Y.~Jiang, Q.~Luo, X.~Ma, P.~W. Chiu, Z.~Li, and
  P.~Kazanzides, ``Flexivision: Teleporting the surgeon’s eyes via robotic
  flexible endoscope and head-mounted display,'' in \emph{2020 IEEE/RSJ
  International Conference on Intelligent Robots and Systems (IROS)}.\hskip 1em
  plus 0.5em minus 0.4em\relax IEEE, 2020, pp. 3281--3287.

\bibitem{gao2023laparoscopic}
S.~Gao, Y.~Wang, H.~Zhou, K.~Yang, Y.~Jiang, L.~Lu, S.~Wang, X.~Ma, B.~C.
  Nephew, L.~Fichera \emph{et~al.}, ``Laparoscopic photoacoustic imaging system
  integrated with the da vinci surgical system,'' in \emph{Medical Imaging
  2023: Image-Guided Procedures, Robotic Interventions, and Modeling}, vol.
  12466.\hskip 1em plus 0.5em minus 0.4em\relax SPIE, 2023, pp. 62--70.

\end{thebibliography}

\end{document}